\pdfoutput=1
\documentclass[11pt]{article}
\usepackage[]{emnlp2021}
\usepackage{times}
\usepackage{latexsym}
\usepackage[T1]{fontenc}
\usepackage[utf8]{inputenc}
\usepackage{microtype}
\usepackage{times,latexsym}
\usepackage{url}
\usepackage[T1]{fontenc}
\usepackage{todonotes}
\usepackage{multirow}
\usepackage{array,booktabs,ragged2e}
\usepackage{multirow}
\usepackage{pifont} 
\usepackage{subcaption}
\usepackage{amsmath, balance}
\usepackage{caption}
\usepackage{subcaption}
\usepackage{booktabs}
\usepackage{soul}

\newcolumntype{R}[1]{>{\RaggedLeft\arraybackslash}p{#1}}
\newcolumntype{L}[1]{>{\RaggedRight\arraybackslash}p{#1}}

\newcommand\blfootnote[1]{%
  \begingroup
  \renewcommand\thefootnote{}\footnote{#1}%
  \addtocounter{footnote}{-1}%
  \endgroup
}

\newcommand{\cmark}{\ding{51}}

\newcommand\coauth{$^\star$}
\newcommand{\gt}{$^\dagger$}
\newcommand{\ucsd}{$^\diamond$}

\title{Latent Hatred: {A} Benchmark for Understanding Implicit Hate Speech} 

\author{Mai ElSherief \coauth \ucsd \hspace{1.5em}
        Caleb Ziems \coauth \gt \hspace{1.5em}
        David Muchlinski\gt \hspace{1.5em}
        \textbf{Vaishnavi Anupindi}\gt\\
        \textbf{Jordyn Seybolt}\gt \hspace{1.5em}
        \textbf{Munmun De Choudhury}\gt \hspace{1.5em}
        \textbf{Diyi Yang}\gt \hspace{1.5em} \\
        \ucsd UC San Diego, \gt Georgia Institute of Technology\\
        \texttt{melsherief@ucsd.edu} \\
        \texttt{\{cziems, dmuchlinski3, vanupindi3\}@gatech.edu} \\
        \texttt{\{jseybolt3, munmund, dyang888\}@gatech.edu}
}

\begin{document}
\maketitle
\begin{abstract}
Hate speech has grown significantly on social media, causing serious consequences for victims of all demographics. Despite much attention being paid to characterize and detect discriminatory speech, most work has focused on explicit or overt hate speech, failing to address a more pervasive form based on coded or indirect language. To fill this gap, this work introduces a theoretically-justified taxonomy of \textit{implicit hate speech} and a benchmark corpus with fine-grained labels for each message and its implication. We present systematic analyses of our dataset using contemporary baselines to detect and explain implicit hate speech, and we discuss key features that challenge existing models. This dataset will continue to serve as a useful benchmark for understanding this multifaceted issue. To download the data, see \url{https://github.com/GT-SALT/implicit-hate}\blfootnote{\coauth Equal contribution.}
\end{abstract}

\section{Introduction}
\begin{figure}[!tb]
\centering
\includegraphics[width=8cm, height = 6cm]{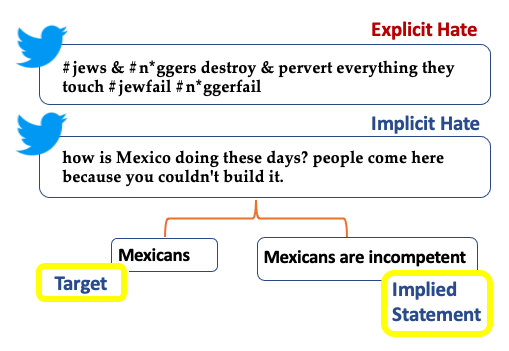}
\caption{Sample posts from our dataset outlining the differences between explicit and implicit hate speech. Explicit hate is \textbf{direct} and leverages specific keywords while implicit hate is more \textbf{abstract}.
Explicit text has been modified to include a star (*).
}
\label{fig:crownjewel}
\end{figure}

Hate speech is pervasive in social media. Platforms have responded by banning hate groups and flagging abusive text \cite{klepper_2020}, and the research community has developed increasingly competitive hate speech detection systems \cite{fortuna2018survey,badjatiya2017deep}. While prior efforts have focused extensively on overt abuse or \textit{explicit hate speech} \cite{schmidt2017survey}, recent works have started to highlight the diverse range of \textit{implicitly} hateful messages that have previously gone unnoticed by moderators and researchers alike \cite{jurgens2019just,waseem2017understanding,qian2019learning}. Figure~\ref{fig:crownjewel} provides an example from each hate speech type (explicit vs. implicit).

Implicit hate speech is defined by \textit{coded} or \textit{indirect} language that disparages a person or group on the basis of protected characteristics like race, gender, and cultural identity~\cite{nockleby2000hate}. Extremist groups have used this coded language to mobilize acts of aggression \cite{gubler2015violent} and domestic terrorism \cite{piazza2020politician} while also maintaining \textit{plausible deniability} for their actions \cite{denigot2020dogwhistles}. Because this speech lacks clear lexical signals, hate groups can evade keyword-based detection systems \cite{waseem2017understanding,wiegand2019detection}, and even the most advanced architectures may suffer if they have not been trained on implicitly abusive messages \cite{caselli2020feel}. 

The primary challenge for statistical and neural classifiers is the linguistic nuance and diversity of the implicit hate class, which includes indirect sarcasm and humor \cite{waseem2016hateful,fortuna2018survey}, euphemisms \cite{magu2018determining}, circumlocution \cite{gao2017detecting}, and other symbolic or metaphorical language \cite{qian2019learning}. The type of implicit hate speech also varies, from dehumanizing comparisons \cite{leader2016dangerous} and stereotypes \cite{warner2012detecting}, to threats, intimidation, and incitement to violence \cite{sanguinetti2018italian,fortuna2018survey}. Importantly, the field lacks a  theoretically-grounded framework and a large-scale dataset to help inform a more empirical understanding of implicit hate in all of its diverse manifestations.

To fill this gap, we establish new resources to sustain research and facilitate both fine-grained classification and generative intervention strategies. Specifically, we develop a 6-class taxonomy of implicit hate speech that is grounded in the social science literature. We use this taxonomy to annotate a new Twitter dataset with broad coverage of the most prevalent hate groups in the United States. This dataset makes three original contributions: (1) it is a large and representative sample of \textit{implicit} hate speech with (2) fine-grained \textit{implicit hate labels} and (3) natural language descriptions of the \textit{implied aspects} for each hateful message. 
Finally, we train competitive baseline classifiers to detect implicit hate speech and generate its implied statements. While state-of-the-art neural models are effective at a high level hate speech classification, they are not effective at spelling out more fine-grained categories with detailed explanations the implied message. The results suggest our dataset can serve as a useful benchmark for understanding implicit hate speech. 
\section{Related Work}  
Numerous hate speech datasets exist, and we summarize them in Table~\ref{tab:hate_speech_data}. The majority are skewed towards \textit{explicitly} abusive text since they were originally seeded with hate lexicons \cite{basile2019semeval,founta2018large,davidson2017automated,waseem2016hateful}, racial identifiers \cite{warner2012detecting}, or explicitly hateful phrases such as ``I hate \textit{<target>}''~\cite{silva2016analyzing}.  Because of a heavy reliance on overt lexical signals,
explicit hate speech datasets have known racial biases \cite{sap2019risk}. Among public datasets, all but one have near or above a 20\% concentration of profanity\footnote{We use the swear word list from https://bit.ly/2SQySZv, excluding ambiguous terms like \textit{bloody}, \textit{prick}, etc.} in the hate class (Table~\ref{tab:hate_speech_data}).

A few neutrally-seeded datasets also exist \cite{burnap2014hate,de2018hate,warner2012detecting}. Although some may contain implicit hate speech, there are no \textit{implicit hate labels} and thus the distribution is unknown. Furthermore, these datasets tend to focus more on controversial events (e.g. the Lee Rigby murder; \citeauthor{burnap2014hate}) or specific hate targets (e.g. immigrants; \citeauthor{basile2019semeval}), which may introduce \textit{topic bias} and artificially inflate model performance on implicit examples \cite{wiegand2019detection}. Consider \citet{sap-etal-2020-social} for example: 31\% of posts take the form of the question leading up to a mean joke.  There is still need for a representative and syntactically diverse implicit hate benchmark.

Our contribution is similar to the Gab Hate Corpus of \citet{kennedy2018gab}, which provides both explicit and implicit hate and target labels for a random sample of 27K Gab messages. We extend this work with a \textit{theoretically-grounded taxonomy} and \textit{fine-grained labels for implicit hate speech} beyond the umbrella categories, Assault on Human Dignity (HD) and Call for Violence (CV). Following the work of  \citet{sap-etal-2020-social}, we provide free-text annotations to capture messages' pragmatic implications. However, we are the first to take this framework, which was originally applied stereotype bias, and extend it to \textit{implicit} hate speech more broadly. Implicitly stereotypical language is just a subset of the implicit hate we cover, since we also include other forms of sarcasm, intimidation or incitement to violence, hidden threats, white grievance, and subtle forms of misinformation. Our work also complements recent efforts to capture and understand \textit{microaggressions} \cite{breitfeller2019finding}, a similarly elusive class that draws on subtle and \textit{unconscious} linguistic reflections of social bias, prejudice and inequality \cite{sue2010microaggressions}. 
Similar to \citet{breitfeller2019finding}, we provide a representative and domain-general typology and dataset, but ours are more representative of \textit{active hate groups} in the United States, and our definitions extend to \textit{intentionally} veiled acts of intimidation, threats, and abuse.

\begin{table*}[t]
\centering
\resizebox{\textwidth}{!}{%
    \begin{tabular}{L{35mm}L{18mm}L{30mm}rrccccc} 
    \toprule
    \textbf{Work} &
    \textbf{Source}&\textbf{Domain / Scope}&\textbf{Size}&\textbf{Balance}&\textbf{Expletives}&\textbf{Public}&\textbf{Target}&\textbf{Implicit}&\textbf{Implied}\\ \midrule
    \citet{basile2019semeval} & Twitter & \href{https://github.com/msang/hateval/blob/master/keyword_set.md}{Misogynistic, \newline anti-immigrant} &  19,600 & Unknown & Unknown &
    \href{https://competitions.codalab.org/competitions/19935}{\cmark} & \cmark & & \\ \hline 
    \citet{burnap2014hate} & Twitter & Lee Rigby murder & 1,901 & 11.7\% & Unknown & & & \\ \hline
    \citet{davidson2017automated} & Twitter & 
    HateBase terms &  24,802 & 5.0\% & 69.8\% & \href{https://github.com/t-davidson/hate-speech-and-offensive-language/tree/master/data}{\cmark} & & & \\ \hline
    \citet{djuric2015hate} & Yahoo \newline Finance & Unknown & 951,736 & 5.9\% & Unknown & & & \\ \hline 
    \citet{founta2018large} & Twitter & Offensive terms & 80,000 & 7.5\% & 73.9\% & \href{https://github.com/ENCASEH2020/hatespeech-twitter}{\cmark} & & \\ \hline 
    \citet{gao2017detecting} & Fox News \newline Comments & Unknown & 1,528 & 28.5\% & Unknown & & & \\ \hline 
    \citet{de2018hate} & Stormfront & One hate group & 9,916 & 11.3\% & 7.8\% & \href{https://github.com/Vicomtech/hate-speech-dataset}{\cmark} & &
    \\ \hline 
    \citet{kennedy2018gab} & Gab & Random sample & 27,665 & 9.1\% & 28.2\% & \href{https://osf.io/edua3/}{\cmark} & \cmark & \cmark \\ \hline 
    \citet{sap-etal-2020-social} & Compilation & Mixed & 44,671 & 44.8\% & 28.5\% & \href{https://homes.cs.washington.edu/~msap/social-bias-frames/}{\cmark} & \cmark & & \cmark \\ \hline 
    \citet{warner2012detecting} & Yahoo + \newline Web & Anti-semitic & 9,000 & Unknown & Unknown & \cmark & & \\ \hline
    \citet{waseem2016hateful} & Twitter & Sexist, racist terms & 16,914 & 31.7\% & 17.6\% & \href{https://github.com/ZeerakW/hatespeech}{\cmark} & & & \\ \hline 
    \citet{zampieri2019predicting} & Twitter & Political phrases & 14,000 & 32.9\% & Unknown &
    \href{https://competitions.codalab.org/competitions/20011\#learn_the_details}{\cmark} & \cmark & & \\ \hline 
    \textsc{Implicit Hate Corpus} (ours) & Twitter & Hate groups & 22,584 & 39.6\% & 3.2\% & \cmark & \cmark & \cmark & \cmark \\ \hline 
    \bottomrule
    \end{tabular}%
    }
\caption{Summary of English hate speech datasets in terms of \textit{Domain / Scope}, \textit{Size}, hate class \textit{Balance} ratio, the proportion of \textit{Expletives} in the hate class, and the inclusion of \textit{Target} demographic, binary \textit{Implicit} hate speech labels, and \textit{Implied} statement summaries.
Most datasets cover a narrow subset of hate speech like anti-semitism or sexism, and do not include implicit hate labels. Ours is the first to include a fine-grained implicit hate taxonomy. 
}
\label{tab:hate_speech_data}
\end{table*}

\section{Taxonomy of Implicit Hate Speech}
\label{section:theory}

Implicit hate speech is a subclass of hate speech defined by \textit{the use of coded or indirect language} such as sarcasm, metaphor and circumlocution to disparage a protected group or individual, or to convey prejudicial and harmful views about them \cite{gao2017recognizing,waseem2017understanding}.
The NLP community has not yet confronted, in a consistent and unified manner, the multiplicity of subtle challenges that implicit hate presents for online communities. 
To this end, we introduce a new typology for characterizing and detecting different forms of implicit hate, based on social science and relevant NLP literature. 
Our categories are not necessarily mutually exclusive, but they represent principle axes of implicit hate, and while they may not be collectively exhaustive, we find they cover 98.6\% of implicit hate in a representative sample of the most prevalent hate ideologies in the U.S.

\textbf{White Grievance} includes frustration over a minority group's perceived privilege and casting majority groups as the real victims of racism \citep{berbrier2000victim,bloch2020playing}.
This language is linked to extremist behavior and support for violence \citep{miller2020hate}. An example is \textit{Black lives matter and white lives don't? Sounds racist.}

\textbf{Incitement to Violence} 
includes flaunting in-group unity and power or elevating known hate groups and ideologies \cite{somerville2011violence}. 
Phrases like \textit{`white brotherhood} operate in the former manner, while statements like \textit{Hitler was Germany -- Germans shall rise again!} operate in the latter, elevating nationalism and Nazism. Article 20 of the UN International Covenant on Civil and Political Rights \cite{assembly1966international} states that speech which incites violence shall be prohibited by law.

\textbf{Inferiority Language} implies one group or individual is inferior to another \cite{nielsen2002subtle}, and it can include dehumanization (denial of a person's humanity), and toxification (language that compares the target with disease, insects, animals), both of which are early warning signs of genocide \cite{leader2016dangerous,neilsen2015toxification}. Inferiority language is also related to \textit{assaults on human dignity} \cite{kennedy2018gab}, \textit{dominance} \cite{saha2018hateminers}, and \textit{declarations of superiority of the in-group} \cite{fortuna2018survey}.
For example, \textit{It's not a coincidence the best places to live are majority white}.

\textbf{Irony} refers to the use of sarcasm \cite{waseem2016hateful,justo2014extracting}, humor \cite{fortuna2018survey}, and satire \cite{sanguinetti2018italian} to attack or demean a protected class or individual. 
For example, in the context of one hate group, the tweet \textit{Horrors... Disney will be forced into hiring Americans} works to discredit Disney for allegedly hiring only non-citizens or, really, non-whites. Irony is not exempt from our hate speech typology, since it is commonly used by modern online hate groups to mask their hatred and extremism \citep{dreisbach_2021}.

\textbf{Stereotypes and Misinformation} associate a protected class with negative attributes such as crime or terrorism \cite{warner2012detecting,sanguinetti2018italian} as in the rhetorical question, \textit{Can someone tell the black people in Chicago to stop killing one another before it becomes Detroit?} This class also includes misinformation that feeds stereotypes and vice versa, like holocaust denial and other forms of historical negationism \cite{belavusau2017hate,cohen2009holocaust}.

\textbf{Threatening and Intimidation} convey a speaker commitment to a target's pain, injury, damage, loss, or violation of rights. While explicitly violent threats are well-recognized in the hate speech literature \cite{sanguinetti2018italian}, here we highlight threats related to implicit violation of rights and freedoms, removal of opportunities, and more subtle forms of intimidation, such as \textit{All immigration of non-whites should be ended.} 

\section{Data Collection and Annotation}
\label{section:data_collection}

We collect and annotate a benchmark dataset for implicit hate language using our taxonomy. 
Our main source of data uses content published by online hate groups and their followers on Twitter for two reasons.  
First, as modern hate groups have become more active online, they provide an increasingly vivid picture of the more subtle and coded forms of hate that we are interested in. 
Second, the problem of hateful misinformation is compounded on social media platforms like Twitter where around 3 out of 4 users get their news \cite{shearer2017news}. This motivates a representative sample of \textit{online} communication exchanged on \textit{Twitter} between members of the \textit{most prominent U.S hate groups}.

We focus on the eight largest ideological clusters of U.S. hate groups as given by the \citet{splc2019} report. These ideological classes are \textit{Black Separatist} (27.1\%), \textit{White Nationalist} (16.4\%), \textit{Neo-Nazi} (6.2\%), \textit{Anti-Muslim} (8.9\%), \textit{Racist Skinhead} (5.1\%), \textit{Ku Klux Klan} (5.0\%), \textit{Anti-LGBT} (7.4\%), and \textit{Anti-Immigrant} (2.12\%). Detailed background and discussion on each hate ideology can be found at the the SPLC Extremist Files page \cite{splc2020}. 

\subsection{Data Collection and Filtering}\label{sec:data_collection}

We matched all SPLC hate groups with their corresponding Twitter accounts using the account names and bios. Then, for each ideological cluster above, we selected the three hate group accounts with the most followers, since these were likely to be the most visible and engaged. We collected all tweets, retweets, and replies from the timelines of our selected hate groups between January 1, 2015 and December 31, 2017, for a total of 4,748,226 tweets, giving us with an broad sample of hate group activity before many accounts were banned.

Hateful content is semantically diverse, with different hate groups motivated by different ideologies. Seeking a representative sample, we identified group-specific salient content from each ideology by performing part of speech (POS) tagging on each tweet. Then we computed the log odds ratio with informative Dirichlet prior~\cite{monroe2008fightin} for each noun, hashtag, and adjective to identify the top 25 words per ideology. After filtering for tweets that contained one of the salient keywords, we ran the 3-way HateSonar classifier of \citet{davidson2017automated} to remove content that was likely to be explicitly hateful. Specifically, we removed all tweets that were classified as \textit{offensive}, and then ran a final sweep over the \textit{neutral} and \textit{hate} categories, removing tweets that contained any explicit keyword found in NoSwear \cite{noSwearing} or Hatebase \cite{hatebase2020}. 

\subsection{Crowdsourcing and Expert Annotation}
\label{subsection:annotation}
To acquire implicit hate speech labels with two different resolutions, we ran two stages of annotation. First, we collected high-level labels, \textit{explicit hate, implicit hate,} or \textit{not hate}.  
Then, we took a second pass through the implicit hate tweets with expert annotation over the fine-grained implicit hate taxonomy from Section~\ref{section:theory}. 

\subsubsection{Stage 1: High Level Categorization} 
\label{subsec:stage_1_annotation}
Amazon Mechanical Turk (MTurk) annotators 
completed our high-level labeling task.
We provided them with a definition of hate speech \cite{twitterPolicy} and examples of explicit, implicit, and non-hateful content (See Appendix~\ref{appdx:data_collection_details}), and required them to pass a short five-question qualification check for understanding with a score of at least 90\% in accordance with crowdsourcing standards~\cite{sheehan2018crowdsourcing}.  
We paid annotators a fair wage above the federal minimum. 
Three workers labeled each tweet, and they reached majority agreement for 95.3\% of tweets, with perfect agreement on 45.6\% of the data.
The Intraclass Correlation for one-way random effects between $k=118$ raters was $ICC(1, k) = 0.616$, which indicates moderate inter-rater agreement. 
Using the majority vote, we obtained consensus labels for 
19,112 labeled tweets in total: 933 \textit{explicit hate}, 4,909 \textit{implicit hate}, and 13,291 \textit{not hateful} tweets.

\subsubsection{Stage 2: Fine-Grained Implicit Hate}
\label{subsec:stage_2_annotation}
To promote a more nuanced understanding of our 4,909 implicit hate tweets, we labeled them using our fine-grained category definitions in Section~\ref{section:theory}, adding \textit{other} and \textit{not hate} to take care of any other situations. 
Since these fine-grained categories were too subtle for MTurk workers,\footnote{We saw less than 30\% agreement when we ran this task over three batches of around 200 tweets each on MTurk.} we hired three research assistants to be our expert annotators. We trained them over multiple sessions 
by walking them through seven small pilot batches
and resolving disagreements after each test until they reached moderate agreement. On the next round of 150 tweets, their independent annotations reached a Fleiss' Kappa of 0.61. 
Each annotator then continued labeling an independent partition of the data. Halfway through this process, we ran another attention check with 150 tweets and found that agreement remained consistent with a Fleiss' Kappa of 0.55. Finally, after filtering out tweets marked as \textit{not hate}, there were 4,153 labeled implicit hate tweets remaining. The per-category statistics are summarized in the \textit{\# Tweets Pre Expn.} column of Table~\ref{tab:impl_class_distr}.

\subsubsection{Corpus Expansion} 
Extreme class imbalance may challenge implicit hate classifiers.
To address this disparity, we expand the minority classes, both with bootstrapping and out-of-domain samples.

For bootstrapping, we trained a 6-way BERT classifier on the 4,153 implicit hate labels in the manner of Section~\ref{subsection:experimental_settings} and ran it on 364,300 unlabeled tweets from our corpus. Then we randomly sampled 1,800 tweets for each of the three minority classes according to the classifications \textit{inferiority}, \textit{irony}, and \textit{threatening}. Finally, we augmented this expansion with out-of-domain (OOD) samples from \citet{kennedy2018gab} and \citet{sap-etal-2020-social}. By drawing both from OOD and bootstrapped in-domain samples, we sought to balance two key limitations: (1) bootstrapped samples may be inherently easier, while (2) OOD samples contain artifacts that allow models to benefit from spurious correlations. Our expert annotators labeled this data, and by adding the minority labels from this process, we improved the class balance for a total of 6,346 implicit tweets shown in the \textit{\#~Tweets~Post~Expn.} column of Table~\ref{tab:impl_class_distr}.

\begin{table}[t]
    \centering
    \small
    \begin{tabular}{L{1.8cm}R{1.3cm}R{1.4cm}R{1.4cm}}
        \toprule
         \textbf{\newline Label} & \textbf{\# Tweets \newline Pre Expn} & \textbf{\# Tweets \newline Post Expn} & \textbf{\% \newline Post Expn} \\ \midrule
         Grievance & 1,455 & 1,538 & 24.2\% \\
         Incitement & 1,176 & 1,269 & 20.0\%\\
         Inferiority & 241 & 863 & 13.6\%\\
         Irony & 134 & 797 & 12.6\% \\
         Stereotypical & 1,032 & 1,133 & 17.9\% \\
         Threatening & 57 & 666 & 10.5\% \\
         Other & 58 & 80 & 1.2\% \\ \midrule
         Total & 4,153 & 6,346 & 100\% \\
         \bottomrule
    \end{tabular}
    \caption{Implicit hate category label distribution before and after the expansion stage}
    \label{tab:impl_class_distr}
\end{table}

\subsubsection{Hate Targets and Implied Statement}
\label{section:implied_annotation}
For each of the 6,346 implicit hate tweets, two separate annotators provided us with the message's \textit{target demographic group} and its \textit{implied statement} in free-text format. Implied statements were formatted as Hearst-like patterns~\cite{indurkhya2010handbook} of the form \textit{<target> \{do, are, commit\} <predicate>}, where \textit{<target>} might be phrases such as \textit{immigrants, black folks}.

\begin{table*}[t]
    \centering
    \scriptsize
    \resizebox{1.0\textwidth}{!}{%
        \begin{tabular}{lccccccccc}
        \toprule
         & \multicolumn{4}{c}{\textbf{Binary Classification}} & \phantom{abc} & \multicolumn{4}{c}{\textbf{Implicit Hate Categories}} \\
         \cmidrule{2-5} \cmidrule{7-10}
        \textbf{Models} & P & R & F & Acc & & P & R & F & Acc \\ \midrule
        Hate Sonar & 39.9 & 48.6 & 43.8 & 54.6 && - & - & - & - \\ 
        Perspective API & 50.1 & 61.3 & 55.2 & 63.7 && - & - & - & - \\
        \midrule 
        SVM (n-grams) & 61.4 & 67.7 & 64.4 & 72.7 && 48.8 & 49.2 & 48.4 & 54.2 \\
        SVM (TF-IDF) & 59.5 & 68.8 & 63.9 & 71.6  && 53.0 & 51.7 & 51.5 & 56.5 \\
        SVM (GloVe) & 56.5 & 65.3 & 60.6 & 69.0 && 46.8 & 48.9 & 46.3 & 51.3 \\
        BERT & \textbf{72.1} & 66.0 & 68.9 & \textbf{78.3} && \textbf{59.1} & 57.9 & 58.0 & 62.9 \\
        BERT + Aug & 67.8 & \textbf{73.2} & \textbf{70.4} & 77.5 && 58.6 & \textbf{59.1} & \textbf{58.6} & \textbf{63.8} \\
        BERT + Aug + Wikidata & 67.6 & 72.3 & 69.9 & 77.3 && 53.9 & 55.3 & 54.4 & 62.8 \\
        BERT + Aug + ConceptNet & 68.6 & 70.0 & 69.3 & 77.4 && 54.0 & 55.4 & 54.3 & 62.5 \\ 
        \bottomrule
    \end{tabular}
    }
    \caption{Classification performance metrics averaged over five random seeds. (\textit{Left}) \textbf{Binary Classification}. Performance metrics for implicit hate vs. not hate classification. (\textit{Right}) \textbf{Implicit Hate Categories}. Macro performance metrics for \textit{fine-grained category} classification via implicit hate taxonomy. Best performance is bolded.
    }
    \label{tab:results_macro}
\end{table*}

\section{Implicit Hate Speech Classification}
We experiment with two classification tasks: (1) distinguishing implicit hate speech from non-hate, and (2) categorizing implicit hate speech using one of the 6 classes in our fine-grained taxonomy. 

\subsection{Experimental Setup}
\label{subsection:experimental_settings}
Using a 60-20-20 split for each task, we trained, validated, and tested SVM and BERT baselines. We tried standard unigrams, TF-IDF, and Glove embedding \cite{pennington2014glove} features and tuned linear SVMs with $C \in \{0.1, 1, 10, 100, 1000\}$. Next, we fine-tuned BERT with the learning rate in \{2e-5, 3e-5, 5e-5\}
and the number of epochs in $\{1, 2, 3, 4\}$.\footnote{We kept $\epsilon=1.0 \times10^{-8}$ and the batch size fixed at 8} 
We also balanced the training data (\textbf{BERT + Aug}) with back-translation from Russian via FairSeq \cite{gehring2017convs2s}, using a grid search over the sampling temperature in \{0.5, 0.7, 0.9\}. Finally, we supplemented the previous methods with knowledge-based features to learn implicit associations between entities.
In detail, we matched tweets to entities like \emph{white people}, \emph{Islam}, and \emph{antifa} from Wikidata Knowledge Graph ~\cite{vrandevcic2014wikidata} (\textbf{BERT + Aug + Wikidata}) and ConceptNet numberbatch~\cite{speer2016conceptnet} (\textbf{BERT + Aug + ConceptNet}) by string-matching unigrams, bigrams, and trigrams. Then we averaged across the pre-trained entity embeddings matched for each message.\footnote{11,163 / 22,584 tweets $(\approx54\%)$ were matched to one Wikidata entity (none were matched to more than one); 22,554 / 22,584 tweets $(>99\%)$ were matched to at least one ConceptNet entity, and the average number of matches per tweet was 14.} 
Finally, we concatenated the 768-dimensional BERT final layer with the 200-dimensional Wikidata (or 300-dimensional ConceptNet) embeddings, and fed this representation into an MLP with two hidden layers of dimension 100 and ReLU activation between them, using categorical Cross Entropy loss.

\subsection{Implicit Hate Classification Results}
\label{subsection:experimental_results}
In binary implicit hate speech classification on the left side of Table~\ref{tab:results_macro}, baseline SVM models offer competitive performance with $F_1$ scores up to 64.4, while the fine-tuned neural models gain up to 6 additional points. The BERT-base model achieves significantly better macro precision than the linear SVMs (72.1 vs. at most 61.4), demonstrating a compositional understanding beyond simple keyword-matching. When we look at our best \textbf{BERT + Aug} model, the implicit category most confused with non-hate was \textit{Incitement} (36.3\% of testing examples were classified as not hate), followed by \textit{White Grievance} (29.6\%), \textit{Stereotypical} (23.3\%), \textit{Inferiority} (12.3\%), \textit{Irony} (9.3\%), and \textit{Threatening} (5.5\%). In our 6-way classification task on the right of Table~\ref{tab:results_macro}, we find that the BERT-base models again outperform the linear models. Augmentation does not significantly improve performance in either task since our data is already well-balanced and representative. 
Interestingly, integrating Wikidata and ConceptNet did not lead to any performance boost either. This suggests detecting implicit hate speech might require more compositional reasoning over the involved entities and we urge future work to investigate this.  
For additional comparisons, we consider a zero-shot setting where we test Google's Perspective API\footnote{\url{https://www.perspectiveapi.com/}} and the HateSonar classifier of \citet{davidson2017automated}. Our fine-tuned baselines significantly outperform both zero-shot baselines, which were trained on explicit hate.

\begin{table*}[!tbh]
    \centering
    \resizebox{\textwidth}{!}{%
        \begin{tabular}{rccccccccc}
        \toprule
         & \multicolumn{4}{c}{\textbf{Target Group}} & \phantom{abc} & \multicolumn{4}{c}{\textbf{Implied Statement}} \\
         \cmidrule{2-5} \cmidrule{7-10}
        \textbf{Models} & BLEU & BLEU$^*$ & Rouge-L  & Rouge-L$^*$ & \phantom{abc}  & BLEU  & BLEU$^*$ & Rouge-L  & Rouge-L$^*$\\
        GPT-gdy & 43.7 & 65.2& 42.9 &63.3 &\phantom{abc} & 41.1 & 58.2& 31 &  45.3\\
        GPT-top-p& 57.7 & 76.8& 55.8 &74.6 &\phantom{abc} & 55.2 & 69.4& 40 & 53.9\\
        GPT-beam & 59.3 & 81 & 57.3 &78.6 &\phantom{abc} & 57.8 & 73.8 & 46.5 & 63.4\\
        \hline
        GPT-2-gdy & 45.3 & 67.6& 44.6 &66 &\phantom{abc} & 42.3 & 59.3& 32.7 & 47.4\\
        GPT-2-top-p& 58.0 & 76.9& 56.2 &74.8 &\phantom{abc} & 55.1 & 69.3& 39.6 & 53.1\\
        GPT-2-beam & \textbf{61.3} & \textbf{83.9}& \textbf{59.6} &\textbf{81.8} &\phantom{abc} & \textbf{58.9} & \textbf{75.3}& \textbf{48.3} & \textbf{65.9}\\
        \bottomrule
    \end{tabular}
    }
    \caption{Evaluation of the generation models for Target Group and Implied Statement. 
    (*) denotes the maximum versus the average score (without asterisk).
    gdy: greedy decoding, beam: beam search with 3 hypotheses, and top-p:  nucleus sampling with $p = 0.92$}
    \label{tab:generation_metrics}
\end{table*}

\subsection{Challenges in Detecting Implicit Hate}

To further understand the challenges of implicit hate detection and promising directions for future work, we investigated 100 randomly sampled false negative errors from our best model in the binary task (BERT+Aug) and found a set of linguistic classes it struggles with.\footnote{For robustness check, we also labeled 100 false positives from the BERT base model and found the distribution of errors remains similar.} 
\textbf{(1) Coded hate symbols} \cite{qian2019learning} such as  \textit{\#WPWW} (white pride world wide), \textit{\#NationalSocialism} (Nazism), and \textit{(((they)))} (an anti-Semitic symbol) are contained in 15\% of instances, and our models fail to grasp their semantics.
While individual sentences appear harmless, implicit hate can occur in \textbf{(2) discourse relations} \cite{de2018hate} (19\% of instances) like the implied causal relation between the conjunction \textit{I like him} and \textit{he's white}. Additionally, misinformation \cite{islam2020deep} and out-group \textbf{(3) entity framing} \cite{phadke2020many} (25\%) can be context-sensitive, as in the headline \textit{three Muslims convicted}. Even positive framing of a negative entity can be problematic, like describing a Nazi soldier as \textit{super cool}.

Inferiority statements like \textit{POC need us and not the other way around} also require a deep understanding of \textbf{(4) commonsense} (11\%) surrounding social norms (e.g. \textit{a dependant is inferior to a supplier}) \cite{forbes2020social}. Other challenge cases contain highly \textbf{(5) metaphorical language} (7\%), like the animal metaphor in \textit{a world without white people : a visual look at a mongrel future}. \textbf{(6) Colloquial or idiomatic speech} (17\%) appears in subtle phrases like \textit{infrastructure is the white man's game}, and \textbf{(7) Irony} (15\%) detection \cite{waseem2016hateful} may require pragmatic reasoning and understanding, such as in the phrase \textit{hey kids, wanna replace white people.} 

When we sample false positives, we find our models are prone to \textbf{(8) identity term bias} \cite{dixon2018measuring}. Given the high density of identity terms like \textit{Jew} and \textit{Black} in hateful contexts, our models overclassified tweets with these terms as \textit{hateful}, and particularly \textit{stereotypical} speech. In a similar manner, our model also incorrectly associated \textit{white grievance} with all \textit{diversity}-related discourse, \textit{incitement} with controversial topics like \textit{war} and \textit{race}, and \textit{inferiority language} with value-laden terms like \textit{valid} and \textit{wealth}. 

To sum up, our dataset contains rich linguistic phenomena and an array of subtleties that challenge current state-of-the-art baselines, which can serve as a useful benchmark and offer multiple new directions for future work.

\section{Explaining Implicit Hate Speech}

This section presents our generation results for natural language explanations of both 
(1) who is being targeted and (2) what the implied message is for each implicitly hateful tweet. Generating such explanations can help content moderators better understand the severity and nature of automatically-flagged messages. Additionally, we echo efforts from social media companies (e.g., Instagram~\cite{bryant2019instagram}) where the application alerts the user when the post is flagged ``offensive,'' and asks them if they really want to post it. This strategy has proven successful in deterring hurtful comments. Our work could inspire a similar strategy for implicit hate speech. By showing the user the implied meaning of their post before it is posted, we would enable them to recognize the severity of their words and possibly reconsider their decision to post. 

\begin{table*}[t]
\resizebox{\textwidth}{!}{
\begin{tabular}{llll}
\toprule
\textbf{Post} & \textbf{Type}  & \textbf{Target Group}  & \textbf{Implied Statement}                                                                          \\ \midrule
\multirow{2}{*}{(a) white revolution is the only solution}   & \multirow{2}{*}{incitement}    & non-white people & non-white people are inferior \\                                                                   &                                & \textit{white people}  & \textit{white people are superior to other races}                                                   \\ \midrule
\multirow{2}{*}{\begin{tabular}[c]{@{}l@{}}(b) yes you are fine in a white majority country. how is  mexico doing these days? people \\ come here  because you couldn't build it.\end{tabular}} & \multirow{3}{*}{inferiority}   & mexican people                 & \begin{tabular}[c]{@{}l@{}}mexican people do not build  things 
\end{tabular}                                                                            \\ 
                                                                                                                                                                                                       &                                & \textit{mexicans}         & \textit{mexicans are incompetent}                                                                   \\ \midrule

\multirow{3}{*}{\begin{tabular}[c]{@{}l@{}}(c) black people only seem to thrive in white \\ countries. how strange\end{tabular}}                                                                       & \multirow{3}{*}{irony}         & black people              & \begin{tabular}[c]{@{}l@{}}black people are inferior 
\end{tabular}     \\
                    &                                & \textit{black people}     & \textit{\begin{tabular}[c]{@{}l@{}}black people can't survive without \\ white people\end{tabular}} \\ \midrule
\multirow{2}{*}{\begin{tabular}[c]{@{}l@{}}(d) soon white people will not be a majority \\ anywhere. you believe\end{tabular}}                                                                        & \multirow{2}{*}{grievance}   & whites               & whites have been driven out of the country                                                                            \\
                                                                                                                                                                                                       &                                & \textit{minorities}       & \textit{minorities want to take over the world}                                                     \\ \midrule
\multirow{2}{*}{\begin{tabular}[c]{@{}l@{}}(e) muslims are always wanting to kill \\ someone! its their duty to kill the infidel.\end{tabular}}                                                        & \multirow{2}{*}{stereotypical} & muslims                   & muslims are killers                                                                                 \\ 
                                                                                                                                                                                                       &                                & \textit{muslims}          & \textit{muslims are violent}                                                                        \\ \midrule
\multirow{2}{*}{\begin{tabular}[c]{@{}l@{}}(f) oh how i wish a man with a gun was \\there! wake up white people! they hate us\end{tabular}}                                                   & \multirow{2}{*}{threatening}   & white people              & white people are hated                                                                              \\ 
                                                                                                                                                                                                       &                                & \textit{non-whites} & \textit{non-whites hate whites}                                                                      \\ \bottomrule
\end{tabular}}
\caption{Example posts from our dataset along with their implicit category labels, the GPT-2 generated target and implied statements (first row of each block),
and the ground truth target and implied statements (final row of each block, in \textit{italics}). Generated implied statements are semantically similar to the ground truth statements.} \label{tab:posts_generation_examples}
\end{table*}
 
\subsection{Task Formulation}
Our goal is to develop a natural language system that,
given a post, generates a hateful post's intended target and hidden implied meanings. Therefore, we formulate the problem as a conditional generation task (i.e., conditioned on the post content). During training, the generation model takes a sequence of tokens as input:

\begin{align*}
\begin{split}
    \textbf{x} = & \{[STR], t_1, t_2, ....., t_n, [SEP], \\ 
    & t_{[G1]}, t_{[G2]}, ...., [SEP], t_{[S1]}, t_{[S2]}, ...., [END]\}
\end{split}
\end{align*}
with start token [STR], tweet tokens $t_1:t_n$, target group $t_{[Gi]}$, and implied statement $t_{[Si]}$, and minimizes the cross-entropy loss $-\sum_l \log P(\tilde{t}_l|t_{<l})$. 

During inference, our goal is to mimic real-world scenarios when only the post is available. Therefore, the input to the model only contains post tokens $t_1:t_n$ and we experiment with multiple decoding strategies: greedy search (gdy), beam search, and top-p (nucleus) sampling to generate the explanations $t_{[G_i]}$ and $t_{[S_i]}$. 
 
\subsection{Experiment Setup} 
Our ground-truth comes from the free-text \textit{target demographic} and \textit{implied statement} annotations that we collected for all 6,346 implicit hate tweets in Section~\ref{section:implied_annotation}, with 75\% for training, 12.5\% for validation, and 12.5\% for testing. Since we collect multiple annotations for each post (2 per tweet), we ensure that each post and its corresponding annotations belongs only to one split.

Following recent work on social bias inference and commonsense reasoning \cite{sap-etal-2020-social,forbes2020social,sharma2020computational,bosselut2019comet}, we fine-tune Open-AI's GPT~\cite{radford2018improving} and GPT-2~\cite{radford2019language} pre-trained language models to the task and evaluate using BLEU \cite{papineni2002bleu} and ROUGE-L \cite{lin2004rouge}.

We pick BLEU since it is standard for evaluating machine translation models and ROUGE which is used in summarization contexts; both have been adopted extensively in prior literature.
These automatic metrics indicate the quality of the generated target group and implied statement compared to our annotated ground-truth in terms of n-grams and the longest common sequence overlaps.
Since there are two ground truth annotations per tweet, we measure both the averaged metrics across both references, and the maximum metrics (BLEU$^*$ and ROUGE-L$^*$). 

We tuned hyperparameters and selected the best models based on their performance on the development set, and we reported evaluation results on the test.\footnote{We fine-tune for $e \in {\{1, 2, 3, 5\}}$ epochs with a batch size of 2 and learning rate of $5 \times 10^{-5}$ with linear warm up} For decoding, we generate one frame for greedy decoding and three hypotheses for beam search and top-p (nucleus) sampling with $p=0.92$ and choose the highest scoring frame. 

\subsection{Generation Results}
In Table~\ref{tab:generation_metrics} we find that, GPT-2 outperforms GPT in both \textit{target group} and \textit{implied statement} generation. This difference is likely because GPT-2 was trained on English web text while GPT was trained on fiction books and web text is more similar to our domain.
The BLEU and ROUGE-L scores are higher for the \textit{target group} (e.g., 83.9 BLEU) than for the \textit{implied statement} (e.g., 75.3 BLEU), consistently across both averaged and maximum scores. This is likely because the implied statement is longer, more nuanced, and less likely to be contained in the text itself. Additionally, beam search achieves the highest performance for both GPT and GPT-2, followed by top-p. This is not surprising since both decoding strategies consider multiple hypotheses.
Since BLEU and ROUGE-L measure word overlap and not semantics, it is possible that the results in Table~\ref{tab:generation_metrics} are overly pessimistic. The GPT-2 generated implied statements in Table~\ref{tab:generation_metrics} actually describe the complement (a,d), generalization (b), extrapolation (c), or paraphrase (e,f) of the ground truth, and are thus aligned, despite differences in word choice. 
Overall, our generation results are promising. Transformer-based models may play a key role in explaining the severity and nature of online implicit hate.

\section{Conclusion}
In this work, we introduce a theoretical taxonomy of implicit hate speech and a large-scale benchmark corpus with fine-grained labels for each message and its implication.  
As an initial effort, our work enables the NLP communities to better understand and model implicit hate speech at scale.
We also provide several state-of-the-art baselines for detecting and explaining implicit hate speech.
Experimental results show these neural models can effectively categorize hate speech and spell out more fine-grained implicit hate speech and explaining these hateful messages. 

Additionally, we identified eight challenges in implicit hate speech detection: coded hate symbols, discourse relations, entity framing, commonsense, metaphorical language, colloquial speech, irony, and identity term bias.
To mitigate these challenges, future work could explore deciphering models for coded language~\cite{kambhatla2018decipherment,qian2019learning}, lifelong learning of hateful language~\cite{qian2021lifelong}, contextualized sarcasm detection, and bias mitigation for named entities in hate speech detection systems~\cite{xia2020demoting} and their connection with our dataset.

We demonstrate that our corpus can serve as a useful research benchmark for understanding implicit hate speech online. 
Our work also has implications towards the emerging directions of countering online hate speech~\cite{citron2011intermediaries,mathew2019thou}, detecting online radicalization~\cite{ferrara2016predicting} and modeling societal systematic racism, prejudicial expressions, and biases~\cite{davidson-etal-2019-racial,manzini-etal-2019-black,blodgett2020language}.

\section*{Ethical Considerations}
This study has been approved by the Institutional Review Board (IRB)   at the researchers' institution. For the annotation process, we included a warning in the instructions that the content might be offensive or upsetting. Annotators were also encouraged to stop the labeling process if they were overwhelmed. We also acknowledge the risk associated with releasing an implicit hate dataset. However, we believe that the benefit of shedding light on the implicit hate phenomenon outweighs any risks associated with the dataset release.

\section*{Acknowledgments}
The authors would like to thank the members of SALT lab and the anonymous reviewers for their thoughtful feedback. The annotation process was funded through the School of Interactive Computing at Georgia Tech and the Institute for Data Engineering and Science (IDEAS) Data Curation Award to ElSherief and Yang. The work is supported in part by Russell Sage Foundation. 

\bibliographystyle{acl_natbib}

\begin{thebibliography}{78}
\expandafter\ifx\csname natexlab\endcsname\relax\def\natexlab#1{#1}\fi

\bibitem[{Assembly(1966)}]{assembly1966international}
UN~General Assembly. 1966.
\newblock International covenant on civil and political rights.
\newblock \emph{United Nations, Treaty Series}, 999:171.

\bibitem[{Badjatiya et~al.(2017)Badjatiya, Gupta, Gupta, and
  Varma}]{badjatiya2017deep}
Pinkesh Badjatiya, Shashank Gupta, Manish Gupta, and Vasudeva Varma. 2017.
\newblock Deep learning for hate speech detection in tweets.
\newblock In \emph{Proceedings of the 26th International Conference on World
  Wide Web Companion}, pages 759--760.

\bibitem[{Basile et~al.(2019)Basile, Bosco, Fersini, Nozza, Patti,
  Rangel~Pardo, Rosso, and Sanguinetti}]{basile2019semeval}
Valerio Basile, Cristina Bosco, Elisabetta Fersini, Debora Nozza, Viviana
  Patti, Francisco~Manuel Rangel~Pardo, Paolo Rosso, and Manuela Sanguinetti.
  2019.
\newblock \href {https://doi.org/10.18653/v1/S19-2007} {{S}em{E}val-2019 task
  5: Multilingual detection of hate speech against immigrants and women in
  {T}witter}.
\newblock In \emph{Proceedings of the 13th International Workshop on Semantic
  Evaluation}, pages 54--63, Minneapolis, Minnesota, USA. Association for
  Computational Linguistics.

\bibitem[{Belavusau(2017)}]{belavusau2017hate}
Uladzislau Belavusau. 2017.
\newblock Hate speech.
\newblock \emph{Max Planck Encyclopedia of Comparative Constitutional Law
  (Oxford University Press, 2017 Forthcoming)}.

\bibitem[{Berbrier(2000)}]{berbrier2000victim}
Mitch Berbrier. 2000.
\newblock The victim ideology of white supremacists and white separatists in
  the united states.
\newblock \emph{Sociological Focus}, 33(2):175--191.

\bibitem[{Bloch et~al.(2020)Bloch, Taylor, and Martinez}]{bloch2020playing}
Katrina~Rebecca Bloch, Tiffany Taylor, and Karen Martinez. 2020.
\newblock Playing the race card: White injury, white victimhood and the paradox
  of colour-blind ideology in anti-immigrant discourse.
\newblock \emph{Ethnic and Racial Studies}, 43(7):1130--1148.

\bibitem[{Blodgett et~al.(2020)Blodgett, Barocas, Daum{\'e}~III, and
  Wallach}]{blodgett2020language}
Su~Lin Blodgett, Solon Barocas, Hal Daum{\'e}~III, and Hanna Wallach. 2020.
\newblock \href {https://doi.org/10.18653/v1/2020.acl-main.485} {Language
  (technology) is power: A critical survey of {``}bias{''} in {NLP}}.
\newblock In \emph{Proceedings of the 58th Annual Meeting of the Association
  for Computational Linguistics}, pages 5454--5476, Online. Association for
  Computational Linguistics.

\bibitem[{Bosselut et~al.(2019)Bosselut, Rashkin, Sap, Malaviya, Celikyilmaz,
  and Choi}]{bosselut2019comet}
Antoine Bosselut, Hannah Rashkin, Maarten Sap, Chaitanya Malaviya, Asli
  Celikyilmaz, and Yejin Choi. 2019.
\newblock \href {https://doi.org/10.18653/v1/P19-1470} {{COMET}: Commonsense
  transformers for automatic knowledge graph construction}.
\newblock In \emph{Proceedings of the 57th Annual Meeting of the Association
  for Computational Linguistics}, pages 4762--4779, Florence, Italy.
  Association for Computational Linguistics.

\bibitem[{Breitfeller et~al.(2019)Breitfeller, Ahn, Jurgens, and
  Tsvetkov}]{breitfeller2019finding}
Luke Breitfeller, Emily Ahn, David Jurgens, and Yulia Tsvetkov. 2019.
\newblock \href {https://doi.org/10.18653/v1/D19-1176} {Finding
  microaggressions in the wild: A case for locating elusive phenomena in social
  media posts}.
\newblock In \emph{Proceedings of the 2019 Conference on Empirical Methods in
  Natural Language Processing and the 9th International Joint Conference on
  Natural Language Processing (EMNLP-IJCNLP)}, pages 1664--1674, Hong Kong,
  China. Association for Computational Linguistics.

\bibitem[{Bryant(2019)}]{bryant2019instagram}
Miranda Bryant. 2019.
\newblock \href
  {https://www.theguardian.com/technology/2019/jul/09/instagram-bullying-new-feature-do-you-want-to-post-this}
  {Instagram’s anti-bullying ai asks users: `are you sure you want to post
  this?'}.
\newblock \emph{The Guardian}.

\bibitem[{Burnap and Williams(2014)}]{burnap2014hate}
Peter Burnap and Matthew~Leighton Williams. 2014.
\newblock Hate speech, machine classification and statistical modelling of
  information flows on twitter: Interpretation and communication for policy
  decision making.
\newblock \emph{Pre-print}.

\bibitem[{Caselli et~al.(2020)Caselli, Basile, Mitrovi{\'c}, Kartoziya, and
  Granitzer}]{caselli2020feel}
Tommaso Caselli, Valerio Basile, Jelena Mitrovi{\'c}, Inga Kartoziya, and
  Michael Granitzer. 2020.
\newblock \href {https://aclanthology.org/2020.lrec-1.760} {{I} feel offended,
  don{'}t be abusive! implicit/explicit messages in offensive and abusive
  language}.
\newblock In \emph{Proceedings of the 12th Language Resources and Evaluation
  Conference}, pages 6193--6202, Marseille, France. European Language Resources
  Association.

\bibitem[{Citron and Norton(2011)}]{citron2011intermediaries}
Danielle~Keats Citron and Helen Norton. 2011.
\newblock Intermediaries and hate speech: Fostering digital citizenship for our
  information age.
\newblock \emph{Boston University Law Review}, 91:1435.

\bibitem[{Cohen-Almagor(2009)}]{cohen2009holocaust}
Raphael Cohen-Almagor. 2009.
\newblock Holocaust denial is a form of hate speech.
\newblock In \emph{Amsterdam Law Forum}, volume~2, pages 33--42.

\bibitem[{Davidson et~al.(2019)Davidson, Bhattacharya, and
  Weber}]{davidson-etal-2019-racial}
Thomas Davidson, Debasmita Bhattacharya, and Ingmar Weber. 2019.
\newblock \href {https://doi.org/10.18653/v1/W19-3504} {Racial bias in hate
  speech and abusive language detection datasets}.
\newblock In \emph{Proceedings of the Third Workshop on Abusive Language
  Online}, pages 25--35, Florence, Italy. Association for Computational
  Linguistics.

\bibitem[{Davidson et~al.(2017)Davidson, Warmsley, Macy, and
  Weber}]{davidson2017automated}
Thomas Davidson, Dana Warmsley, Michael Macy, and Ingmar Weber. 2017.
\newblock \href {https://arxiv.org/abs/1703.04009} {Automated hate speech
  detection and the problem of offensive language}.
\newblock \emph{ArXiv preprint}, abs/1703.04009.

\bibitem[{D{\'e}nigot and Burnett(2020)}]{denigot2020dogwhistles}
Quentin D{\'e}nigot and Heather Burnett. 2020.
\newblock \href {https://aclanthology.org/2020.pam-1.3} {Dogwhistles as
  identity-based interpretative variation}.
\newblock In \emph{Proceedings of the Probability and Meaning Conference (PaM
  2020)}, pages 17--25, Gothenburg. Association for Computational Linguistics.

\bibitem[{Dixon et~al.(2018)Dixon, Li, Sorensen, Thain, and
  Vasserman}]{dixon2018measuring}
Lucas Dixon, John Li, Jeffrey Sorensen, Nithum Thain, and Lucy Vasserman. 2018.
\newblock Measuring and mitigating unintended bias in text classification.
\newblock In \emph{Proceedings of the 2018 AAAI/ACM Conference on AI, Ethics,
  and Society}, pages 67--73.

\bibitem[{Djuric et~al.(2015)Djuric, Zhou, Morris, Grbovic, Radosavljevic, and
  Bhamidipati}]{djuric2015hate}
Nemanja Djuric, Jing Zhou, Robin Morris, Mihajlo Grbovic, Vladan Radosavljevic,
  and Narayan Bhamidipati. 2015.
\newblock Hate speech detection with comment embeddings.
\newblock In \emph{Proceedings of the 24th international conference on world
  wide web}, pages 29--30.

\bibitem[{Dreisbach(2021)}]{dreisbach_2021}
Tom Dreisbach. 2021.
\newblock \href
  {https://www.npr.org/2021/04/26/990274685/how-extremists-weaponize-irony-to-spread-hate}
  {How extremists weaponize irony to spread hate}.

\bibitem[{Ferrara et~al.(2016)Ferrara, Wang, Varol, Flammini, and
  Galstyan}]{ferrara2016predicting}
Emilio Ferrara, Wen-Qiang Wang, Onur Varol, Alessandro Flammini, and Aram
  Galstyan. 2016.
\newblock Predicting online extremism, content adopters, and interaction
  reciprocity.
\newblock In \emph{International conference on social informatics}, pages
  22--39. Springer.

\bibitem[{Forbes et~al.(2020)Forbes, Hwang, Shwartz, Sap, and
  Choi}]{forbes2020social}
Maxwell Forbes, Jena~D. Hwang, Vered Shwartz, Maarten Sap, and Yejin Choi.
  2020.
\newblock \href {https://doi.org/10.18653/v1/2020.emnlp-main.48} {Social
  chemistry 101: Learning to reason about social and moral norms}.
\newblock In \emph{Proceedings of the 2020 Conference on Empirical Methods in
  Natural Language Processing (EMNLP)}, pages 653--670, Online. Association for
  Computational Linguistics.

\bibitem[{Fortuna and Nunes(2018)}]{fortuna2018survey}
Paula Fortuna and S{\'e}rgio Nunes. 2018.
\newblock A survey on automatic detection of hate speech in text.
\newblock \emph{ACM Computing Surveys (CSUR)}, 51(4):1--30.

\bibitem[{Founta et~al.(2018)Founta, Djouvas, Chatzakou, Leontiadis, Blackburn,
  Stringhini, Vakali, Sirivianos, and Kourtellis}]{founta2018large}
Antigoni-Maria Founta, Constantinos Djouvas, Despoina Chatzakou, Ilias
  Leontiadis, Jeremy Blackburn, Gianluca Stringhini, Athena Vakali, Michael
  Sirivianos, and Nicolas Kourtellis. 2018.
\newblock Large scale crowdsourcing and characterization of twitter abusive
  behavior.
\newblock \emph{Proceedings of the 12th International AAAI Conference on Web
  and Social Media}.

\bibitem[{Gao and Huang(2017)}]{gao2017detecting}
Lei Gao and Ruihong Huang. 2017.
\newblock \href {https://doi.org/10.26615/978-954-452-049-6_036} {Detecting
  online hate speech using context aware models}.
\newblock In \emph{Proceedings of the International Conference Recent Advances
  in Natural Language Processing, {RANLP} 2017}, pages 260--266, Varna,
  Bulgaria. INCOMA Ltd.

\bibitem[{Gao et~al.(2017)Gao, Kuppersmith, and Huang}]{gao2017recognizing}
Lei Gao, Alexis Kuppersmith, and Ruihong Huang. 2017.
\newblock \href {https://aclanthology.org/I17-1078} {Recognizing explicit and
  implicit hate speech using a weakly supervised two-path bootstrapping
  approach}.
\newblock In \emph{Proceedings of the Eighth International Joint Conference on
  Natural Language Processing (Volume 1: Long Papers)}, pages 774--782, Taipei,
  Taiwan. Asian Federation of Natural Language Processing.

\bibitem[{Gehring et~al.(2017)Gehring, Auli, Grangier, Yarats, and
  Dauphin}]{gehring2017convs2s}
Jonas Gehring, Michael Auli, David Grangier, Denis Yarats, and Yann~N. Dauphin.
  2017.
\newblock \href {http://proceedings.mlr.press/v70/gehring17a.html}
  {Convolutional sequence to sequence learning}.
\newblock In \emph{Proceedings of the 34th International Conference on Machine
  Learning, {ICML} 2017, Sydney, NSW, Australia, 6-11 August 2017}, volume~70
  of \emph{Proceedings of Machine Learning Research}, pages 1243--1252. {PMLR}.

\bibitem[{de~Gibert et~al.(2018)de~Gibert, Perez, Garc{\'\i}a-Pablos, and
  Cuadros}]{de2018hate}
Ona de~Gibert, Naiara Perez, Aitor Garc{\'\i}a-Pablos, and Montse Cuadros.
  2018.
\newblock \href {https://doi.org/10.18653/v1/W18-5102} {Hate speech dataset
  from a white supremacy forum}.
\newblock In \emph{Proceedings of the 2nd Workshop on Abusive Language Online
  ({ALW}2)}, pages 11--20, Brussels, Belgium. Association for Computational
  Linguistics.

\bibitem[{Gubler and Kalmoe(2015)}]{gubler2015violent}
Joshua~R Gubler and Nathan~P Kalmoe. 2015.
\newblock Violent rhetoric in protracted group conflicts: Experimental evidence
  from israel and india.
\newblock \emph{Political Research Quarterly}, 68(4):651--664.

\bibitem[{Hatebase(2020)}]{hatebase2020}
Hatebase. 2020.
\newblock \href {https://hatebase.org/} {[link]}.

\bibitem[{Indurkhya and Damerau(2010)}]{indurkhya2010handbook}
Nitin Indurkhya and Fred~J Damerau. 2010.
\newblock \emph{Handbook of natural language processing}, volume~2.
\newblock CRC Press.

\bibitem[{Islam et~al.(2020)Islam, Liu, Wang, and Xu}]{islam2020deep}
Md~Rafiqul Islam, Shaowu Liu, Xianzhi Wang, and Guandong Xu. 2020.
\newblock Deep learning for misinformation detection on online social networks:
  a survey and new perspectives.
\newblock \emph{Social Network Analysis and Mining}, 10(1):1--20.

\bibitem[{Jones(2020)}]{noSwearing}
Ryan Jones. 2020.
\newblock \href {https://www.noswearing.com/dictionary} {List of swear words,
  bad words, \& curse words - starting with a}.

\bibitem[{Jurgens et~al.(2019)Jurgens, Hemphill, and
  Chandrasekharan}]{jurgens2019just}
David Jurgens, Libby Hemphill, and Eshwar Chandrasekharan. 2019.
\newblock \href {https://doi.org/10.18653/v1/P19-1357} {A just and
  comprehensive strategy for using {NLP} to address online abuse}.
\newblock In \emph{Proceedings of the 57th Annual Meeting of the Association
  for Computational Linguistics}, pages 3658--3666, Florence, Italy.
  Association for Computational Linguistics.

\bibitem[{Justo et~al.(2014)Justo, Corcoran, Lukin, Walker, and
  Torres}]{justo2014extracting}
Raquel Justo, Thomas Corcoran, Stephanie~M Lukin, Marilyn Walker, and
  M~In{\'e}s Torres. 2014.
\newblock Extracting relevant knowledge for the detection of sarcasm and
  nastiness in the social web.
\newblock \emph{Knowledge-Based Systems}, 69:124--133.

\bibitem[{Kambhatla et~al.(2018)Kambhatla, Mansouri~Bigvand, and
  Sarkar}]{kambhatla2018decipherment}
Nishant Kambhatla, Anahita Mansouri~Bigvand, and Anoop Sarkar. 2018.
\newblock \href {https://doi.org/10.18653/v1/D18-1102} {Decipherment of
  substitution ciphers with neural language models}.
\newblock In \emph{Proceedings of the 2018 Conference on Empirical Methods in
  Natural Language Processing}, pages 869--874, Brussels, Belgium. Association
  for Computational Linguistics.

\bibitem[{Kennedy et~al.(2018)Kennedy, Atari, Davani, Yeh, Omrani, Kim,
  Coombs~Jr, Havaldar, Portillo-Wightman, Gonzalez et~al.}]{kennedy2018gab}
Brendan Kennedy, Mohammad Atari, Aida~M Davani, Leigh Yeh, Ali Omrani, Yehsong
  Kim, Kris Coombs~Jr, Shreya Havaldar, Gwenyth Portillo-Wightman, Elaine
  Gonzalez, et~al. 2018.
\newblock The gab hate corpus: A collection of 27k posts annotated for hate
  speech.
\newblock \emph{PsyArXiv. July}, 18.

\bibitem[{Klepper(2020)}]{klepper_2020}
David Klepper. 2020.
\newblock \href
  {https://abcnews.go.com/Business/wireStory/facebook-removes-200-accounts-tied-hate-groups-71101914}
  {Facebook removes nearly 200 accounts tied to hate groups}.

\bibitem[{Leader~Maynard and Benesch(2016)}]{leader2016dangerous}
Jonathan Leader~Maynard and Susan Benesch. 2016.
\newblock Dangerous speech and dangerous ideology: An integrated model for
  monitoring and prevention.
\newblock \emph{Genocide Studies and Prevention}.

\bibitem[{Lin(2004)}]{lin2004rouge}
Chin-Yew Lin. 2004.
\newblock \href {https://aclanthology.org/W04-1013} {{ROUGE}: A package for
  automatic evaluation of summaries}.
\newblock In \emph{Text Summarization Branches Out}, pages 74--81, Barcelona,
  Spain. Association for Computational Linguistics.

\bibitem[{Magu and Luo(2018)}]{magu2018determining}
Rijul Magu and Jiebo Luo. 2018.
\newblock \href {https://doi.org/10.18653/v1/W18-5112} {Determining code words
  in euphemistic hate speech using word embedding networks}.
\newblock In \emph{Proceedings of the 2nd Workshop on Abusive Language Online
  ({ALW}2)}, pages 93--100, Brussels, Belgium. Association for Computational
  Linguistics.

\bibitem[{Manzini et~al.(2019)Manzini, Yao~Chong, Black, and
  Tsvetkov}]{manzini-etal-2019-black}
Thomas Manzini, Lim Yao~Chong, Alan~W Black, and Yulia Tsvetkov. 2019.
\newblock \href {https://doi.org/10.18653/v1/N19-1062} {Black is to criminal as
  caucasian is to police: Detecting and removing multiclass bias in word
  embeddings}.
\newblock In \emph{Proceedings of the 2019 Conference of the North {A}merican
  Chapter of the Association for Computational Linguistics: Human Language
  Technologies, Volume 1 (Long and Short Papers)}, pages 615--621, Minneapolis,
  Minnesota. Association for Computational Linguistics.

\bibitem[{Mathew et~al.(2019)Mathew, Saha, Tharad, Rajgaria, Singhania, Maity,
  Goyal, and Mukherjee}]{mathew2019thou}
Binny Mathew, Punyajoy Saha, Hardik Tharad, Subham Rajgaria, Prajwal Singhania,
  Suman~Kalyan Maity, Pawan Goyal, and Animesh Mukherjee. 2019.
\newblock Thou shalt not hate: Countering online hate speech.
\newblock In \emph{Proceedings of the International AAAI Conference on Web and
  Social Media}, volume~13, pages 369--380.

\bibitem[{Miller-Idriss(2020)}]{miller2020hate}
Cynthia Miller-Idriss. 2020.
\newblock \emph{Hate in the homeland: The new global far right}.
\newblock Princeton University Press.

\bibitem[{Monroe et~al.(2008)Monroe, Colaresi, and Quinn}]{monroe2008fightin}
Burt~L Monroe, Michael~P Colaresi, and Kevin~M Quinn. 2008.
\newblock Fightin'words: Lexical feature selection and evaluation for
  identifying the content of political conflict.
\newblock \emph{Political Analysis}, 16(4):372--403.

\bibitem[{Neilsen(2015)}]{neilsen2015toxification}
Rhiannon~S Neilsen. 2015.
\newblock ‘toxification’as a more precise early warning sign for genocide
  than dehumanization? an emerging research agenda.
\newblock \emph{Genocide Studies and Prevention: An International Journal},
  9(1):9.

\bibitem[{Nielsen(2002)}]{nielsen2002subtle}
Laura~Beth Nielsen. 2002.
\newblock Subtle, pervasive, harmful: Racist and sexist remarks in public as
  hate speech.
\newblock \emph{Journal of Social Issues}, 58(2):265--280.

\bibitem[{Nockleby(2000)}]{nockleby2000hate}
John~T Nockleby. 2000.
\newblock Hate speech.
\newblock \emph{Encyclopedia of the American constitution}, 3(2):1277--1279.

\bibitem[{Papineni et~al.(2002)Papineni, Roukos, Ward, and
  Zhu}]{papineni2002bleu}
Kishore Papineni, Salim Roukos, Todd Ward, and Wei-Jing Zhu. 2002.
\newblock \href {https://doi.org/10.3115/1073083.1073135} {{B}leu: a method for
  automatic evaluation of machine translation}.
\newblock In \emph{Proceedings of the 40th Annual Meeting of the Association
  for Computational Linguistics}, pages 311--318, Philadelphia, Pennsylvania,
  USA. Association for Computational Linguistics.

\bibitem[{Pennington et~al.(2014)Pennington, Socher, and
  Manning}]{pennington2014glove}
Jeffrey Pennington, Richard Socher, and Christopher Manning. 2014.
\newblock \href {https://doi.org/10.3115/v1/D14-1162} {{G}lo{V}e: Global
  vectors for word representation}.
\newblock In \emph{Proceedings of the 2014 Conference on Empirical Methods in
  Natural Language Processing ({EMNLP})}, pages 1532--1543, Doha, Qatar.
  Association for Computational Linguistics.

\bibitem[{Phadke and Mitra(2020)}]{phadke2020many}
Shruti Phadke and Tanushree Mitra. 2020.
\newblock \href {https://doi.org/10.1145/3313831.3376456} {Many faced hate: {A}
  cross platform study of content framing and information sharing by online
  hate groups}.
\newblock In \emph{{CHI} '20: {CHI} Conference on Human Factors in Computing
  Systems, Honolulu, HI, USA, April 25-30, 2020}, pages 1--13. {ACM}.

\bibitem[{Piazza(2020)}]{piazza2020politician}
James~A Piazza. 2020.
\newblock Politician hate speech and domestic terrorism.
\newblock \emph{International Interactions}, pages 1--23.

\bibitem[{Qian et~al.(2019)Qian, ElSherief, Belding, and
  Wang}]{qian2019learning}
Jing Qian, Mai ElSherief, Elizabeth Belding, and William~Yang Wang. 2019.
\newblock \href {https://doi.org/10.18653/v1/N19-1305} {Learning to decipher
  hate symbols}.
\newblock In \emph{Proceedings of the 2019 Conference of the North {A}merican
  Chapter of the Association for Computational Linguistics: Human Language
  Technologies, Volume 1 (Long and Short Papers)}, pages 3006--3015,
  Minneapolis, Minnesota. Association for Computational Linguistics.

\bibitem[{Qian et~al.(2021)Qian, Wang, ElSherief, and Yan}]{qian2021lifelong}
Jing Qian, Hong Wang, Mai ElSherief, and Xifeng Yan. 2021.
\newblock \href {https://doi.org/10.18653/v1/2021.naacl-main.183} {Lifelong
  learning of hate speech classification on social media}.
\newblock In \emph{Proceedings of the 2021 Conference of the North American
  Chapter of the Association for Computational Linguistics: Human Language
  Technologies}, pages 2304--2314, Online. Association for Computational
  Linguistics.

\bibitem[{Radford et~al.(2018)Radford, Narasimhan, Salimans, and
  Sutskever}]{radford2018improving}
Alec Radford, Karthik Narasimhan, Tim Salimans, and Ilya Sutskever. 2018.
\newblock Improving language understanding with unsupervised learning.
\newblock \emph{Technical report, OpenAI}.

\bibitem[{Radford et~al.(2019)Radford, Wu, Child, Luan, Amodei, and
  Sutskever}]{radford2019language}
Alec Radford, Jeffrey Wu, Rewon Child, David Luan, Dario Amodei, and Ilya
  Sutskever. 2019.
\newblock Language models are unsupervised multitask learners.
\newblock \emph{OpenAI blog}, 1(8):9.

\bibitem[{Saha et~al.(2018)Saha, Mathew, Goyal, and
  Mukherjee}]{saha2018hateminers}
Punyajoy Saha, Binny Mathew, Pawan Goyal, and Animesh Mukherjee. 2018.
\newblock \href {https://arxiv.org/abs/1812.06700} {Hateminers: detecting hate
  speech against women}.
\newblock \emph{ArXiv preprint}, abs/1812.06700.

\bibitem[{Sanguinetti et~al.(2018)Sanguinetti, Poletto, Bosco, Patti, and
  Stranisci}]{sanguinetti2018italian}
Manuela Sanguinetti, Fabio Poletto, Cristina Bosco, Viviana Patti, and Marco
  Stranisci. 2018.
\newblock An italian twitter corpus of hate speech against immigrants.
\newblock In \emph{Proceedings of the Eleventh International Conference on
  Language Resources and Evaluation (LREC 2018)}.

\bibitem[{Sap et~al.(2019)Sap, Card, Gabriel, Choi, and Smith}]{sap2019risk}
Maarten Sap, Dallas Card, Saadia Gabriel, Yejin Choi, and Noah~A. Smith. 2019.
\newblock \href {https://doi.org/10.18653/v1/P19-1163} {The risk of racial bias
  in hate speech detection}.
\newblock In \emph{Proceedings of the 57th Annual Meeting of the Association
  for Computational Linguistics}, pages 1668--1678, Florence, Italy.
  Association for Computational Linguistics.

\bibitem[{Sap et~al.(2020)Sap, Gabriel, Qin, Jurafsky, Smith, and
  Choi}]{sap-etal-2020-social}
Maarten Sap, Saadia Gabriel, Lianhui Qin, Dan Jurafsky, Noah~A. Smith, and
  Yejin Choi. 2020.
\newblock \href {https://doi.org/10.18653/v1/2020.acl-main.486} {Social bias
  frames: Reasoning about social and power implications of language}.
\newblock In \emph{Proceedings of the 58th Annual Meeting of the Association
  for Computational Linguistics}, pages 5477--5490, Online. Association for
  Computational Linguistics.

\bibitem[{Schmidt and Wiegand(2017)}]{schmidt2017survey}
Anna Schmidt and Michael Wiegand. 2017.
\newblock \href {https://doi.org/10.18653/v1/W17-1101} {A survey on hate speech
  detection using natural language processing}.
\newblock In \emph{Proceedings of the Fifth International Workshop on Natural
  Language Processing for Social Media}, pages 1--10, Valencia, Spain.
  Association for Computational Linguistics.

\bibitem[{Sharma et~al.(2020)Sharma, Miner, Atkins, and
  Althoff}]{sharma2020computational}
Ashish Sharma, Adam Miner, David Atkins, and Tim Althoff. 2020.
\newblock \href {https://doi.org/10.18653/v1/2020.emnlp-main.425} {A
  computational approach to understanding empathy expressed in text-based
  mental health support}.
\newblock In \emph{Proceedings of the 2020 Conference on Empirical Methods in
  Natural Language Processing (EMNLP)}, pages 5263--5276, Online. Association
  for Computational Linguistics.

\bibitem[{Shearer and Gottfried(2017)}]{shearer2017news}
Elisa Shearer and Jeffrey Gottfried. 2017.
\newblock News use across social media platforms 2017.
\newblock \emph{Pew Research Center}, 7(9).

\bibitem[{Sheehan(2018)}]{sheehan2018crowdsourcing}
Kim~Bartel Sheehan. 2018.
\newblock Crowdsourcing research: data collection with amazon’s mechanical
  turk.
\newblock \emph{Communication Monographs}, 85(1):140--156.

\bibitem[{Silva et~al.(2016)Silva, Mondal, Correa, Benevenuto, and
  Weber}]{silva2016analyzing}
Leandro Silva, Mainack Mondal, Denzil Correa, Fabr{\'\i}cio Benevenuto, and
  Ingmar Weber. 2016.
\newblock Analyzing the targets of hate in online social media.
\newblock In \emph{10th International AAAI Conference on Web and Social Media},
  pages 687--690. AAAI.

\bibitem[{Somerville(2011)}]{somerville2011violence}
Keith Somerville. 2011.
\newblock Violence, hate speech and inflammatory broadcasting in kenya: The
  problems of definition and identification.
\newblock \emph{Ecquid Novi: African Journalism Studies}, 32(1):82--101.

\bibitem[{Speer et~al.(2017)Speer, Chin, and Havasi}]{speer2016conceptnet}
Robyn Speer, Joshua Chin, and Catherine Havasi. 2017.
\newblock \href {http://aaai.org/ocs/index.php/AAAI/AAAI17/paper/view/14972}
  {Conceptnet 5.5: An open multilingual graph of general knowledge}.
\newblock In \emph{Proceedings of the Thirty-First {AAAI} Conference on
  Artificial Intelligence, February 4-9, 2017, San Francisco, California,
  {USA}}, pages 4444--4451. {AAAI} Press.

\bibitem[{SPLC(2019)}]{splc2019}
SPLC. 2019.
\newblock \href {https://www.splcenter.org/hate-map} {Hate map}.

\bibitem[{SPLC(2020)}]{splc2020}
SPLC. 2020.
\newblock \href
  {https://www.splcenter.org/fighting-hate/extremist-files/ideology}
  {Ideologies}.

\bibitem[{Sue(2010)}]{sue2010microaggressions}
Derald~Wing Sue. 2010.
\newblock \emph{Microaggressions in everyday life: Race, gender, and sexual
  orientation}.
\newblock John Wiley \& Sons.

\bibitem[{Twitter(2021)}]{twitterPolicy}
Twitter. 2021.
\newblock \href
  {https://help.twitter.com/en/rules-and-policies/hateful-conduct-policy}
  {Twitter's policy on hateful conduct | twitter help}.

\bibitem[{Vrande{\v{c}}i{\'c} and Kr{\"o}tzsch(2014)}]{vrandevcic2014wikidata}
Denny Vrande{\v{c}}i{\'c} and Markus Kr{\"o}tzsch. 2014.
\newblock Wikidata: a free collaborative knowledgebase.
\newblock \emph{Communications of the ACM}, 57(10):78--85.

\bibitem[{Warner and Hirschberg(2012)}]{warner2012detecting}
William Warner and Julia Hirschberg. 2012.
\newblock \href {https://aclanthology.org/W12-2103} {Detecting hate speech on
  the world wide web}.
\newblock In \emph{Proceedings of the Second Workshop on Language in Social
  Media}, pages 19--26, Montr{\'e}al, Canada. Association for Computational
  Linguistics.

\bibitem[{Waseem et~al.(2017)Waseem, Davidson, Warmsley, and
  Weber}]{waseem2017understanding}
Zeerak Waseem, Thomas Davidson, Dana Warmsley, and Ingmar Weber. 2017.
\newblock \href {https://doi.org/10.18653/v1/W17-3012} {Understanding abuse: A
  typology of abusive language detection subtasks}.
\newblock In \emph{Proceedings of the First Workshop on Abusive Language
  Online}, pages 78--84, Vancouver, BC, Canada. Association for Computational
  Linguistics.

\bibitem[{Waseem and Hovy(2016)}]{waseem2016hateful}
Zeerak Waseem and Dirk Hovy. 2016.
\newblock \href {https://doi.org/10.18653/v1/N16-2013} {Hateful symbols or
  hateful people? predictive features for hate speech detection on {T}witter}.
\newblock In \emph{Proceedings of the {NAACL} Student Research Workshop}, pages
  88--93, San Diego, California. Association for Computational Linguistics.

\bibitem[{Wiegand et~al.(2019)Wiegand, Ruppenhofer, and
  Kleinbauer}]{wiegand2019detection}
Michael Wiegand, Josef Ruppenhofer, and Thomas Kleinbauer. 2019.
\newblock \href {https://doi.org/10.18653/v1/N19-1060} {{D}etection of
  {A}busive {L}anguage: the {P}roblem of {B}iased {D}atasets}.
\newblock In \emph{Proceedings of the 2019 Conference of the North {A}merican
  Chapter of the Association for Computational Linguistics: Human Language
  Technologies, Volume 1 (Long and Short Papers)}, pages 602--608, Minneapolis,
  Minnesota. Association for Computational Linguistics.

\bibitem[{Xia et~al.(2020)Xia, Field, and Tsvetkov}]{xia2020demoting}
Mengzhou Xia, Anjalie Field, and Yulia Tsvetkov. 2020.
\newblock \href {https://doi.org/10.18653/v1/2020.socialnlp-1.2} {Demoting
  racial bias in hate speech detection}.
\newblock In \emph{Proceedings of the Eighth International Workshop on Natural
  Language Processing for Social Media}, pages 7--14, Online. Association for
  Computational Linguistics.

\bibitem[{Zampieri et~al.(2019)Zampieri, Malmasi, Nakov, Rosenthal, Farra, and
  Kumar}]{zampieri2019predicting}
Marcos Zampieri, Shervin Malmasi, Preslav Nakov, Sara Rosenthal, Noura Farra,
  and Ritesh Kumar. 2019.
\newblock \href {https://doi.org/10.18653/v1/N19-1144} {Predicting the type and
  target of offensive posts in social media}.
\newblock In \emph{Proceedings of the 2019 Conference of the North {A}merican
  Chapter of the Association for Computational Linguistics: Human Language
  Technologies, Volume 1 (Long and Short Papers)}, pages 1415--1420,
  Minneapolis, Minnesota. Association for Computational Linguistics.

\end{thebibliography}

\appendix

\begin{figure*}[htbp]
\centering
\includegraphics[width =\textwidth]{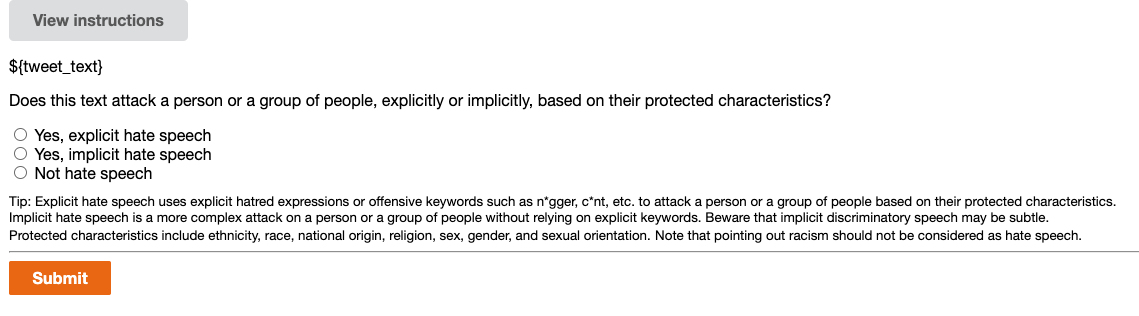}
\caption{Amazon Mechanical Turk interface used to collect ternary annotations (explicit hate, implicit hate, and not hate) for our first stage.} 
\label{fig:amt_stg_1_ui}
\end{figure*}

\section{Data Collection Details}
\label{appdx:data_collection_details}
\begin{figure*}
\begin{subfigure}{\textwidth}
  \centering
  \includegraphics[width=0.9\textwidth]{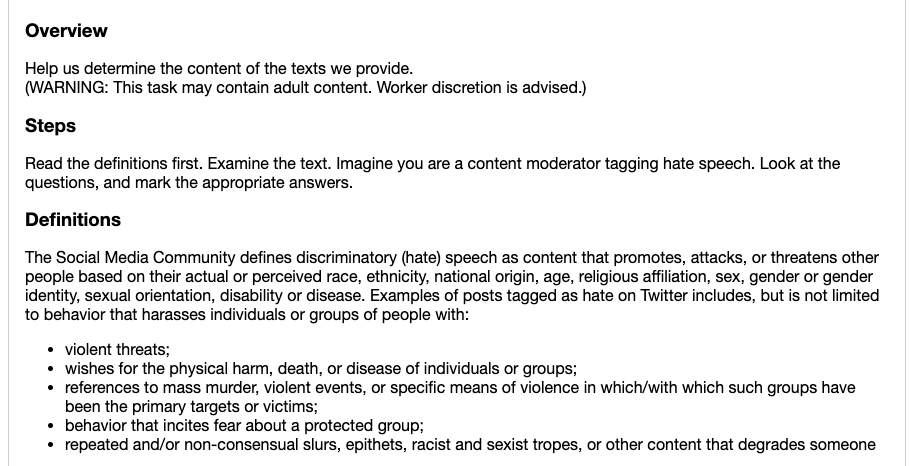}
  \label{fig:sfig1}
\end{subfigure}\\
\begin{subfigure}{\textwidth}
  \centering
  \includegraphics[width=.9\linewidth]{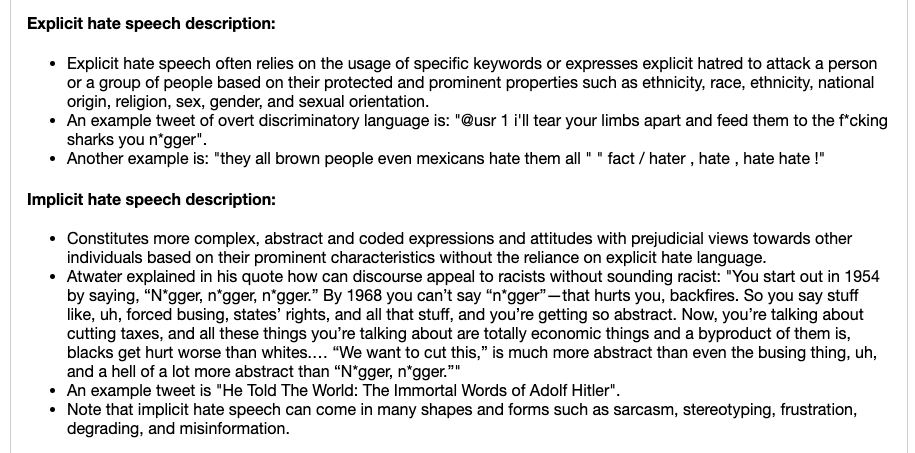}
  \label{fig:sfig2}
\end{subfigure}
\begin{subfigure}{\textwidth}
  \centering
  \includegraphics[width=.9\linewidth]{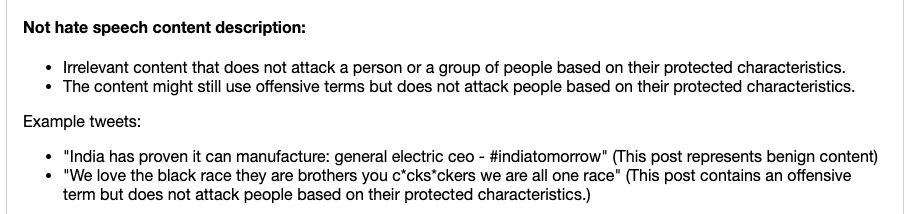}
  \label{fig:sfig3}
\end{subfigure}
\caption{Instructions and examples provided to Amazon Mechanical Turk workers. Our definition of hate speech is grounded in social media communities' rules.}
\label{fig:amt_instructions}
\end{figure*}

In our first annotation stage (Section~\ref{subsec:stage_1_annotation}), we provide a broad definition of hate speech grounded in Twitter’s hateful conduct policy \cite{twitterPolicy}, and detailed definitions for what constitutes explicit hate, implicit hate, and non-hateful content with examples from each class. We explain that explicit hate speech contains explicit keywords directed towards a protected entity. We define implicit hate speech as outlined in the paper and ground this definition in a quote from Lee Atwater on how discourse can appeal to racists without sounding racist: ``\textit{You start out in 1954 by saying, “N*gger, n*gger, n*gger.” By 1968 you can’t say “n*gger”—that hurts you, backfires. So you say stuff like, uh, forced busing, states’ rights, and all that stuff, and you’re getting so abstract}''. To ensure quality, we chose only AMT Master workers who (1) have approval rate >98\% and more than 5000 HITs approved, (2) scored $\geq90\%$ on our five-question qualification test where they must (a) identify the differences between explicit and implicit hate speech and (b) identify the hate target even if the target is not explicitly mentioned. Figures~\ref{fig:amt_stg_1_ui} and~\ref{fig:amt_instructions} depict snippets of the first stage annotation task and the instructions provided to guide the annotators, respectively.

\begin{figure*}[htbp]
\centering
\includegraphics[width =\textwidth]{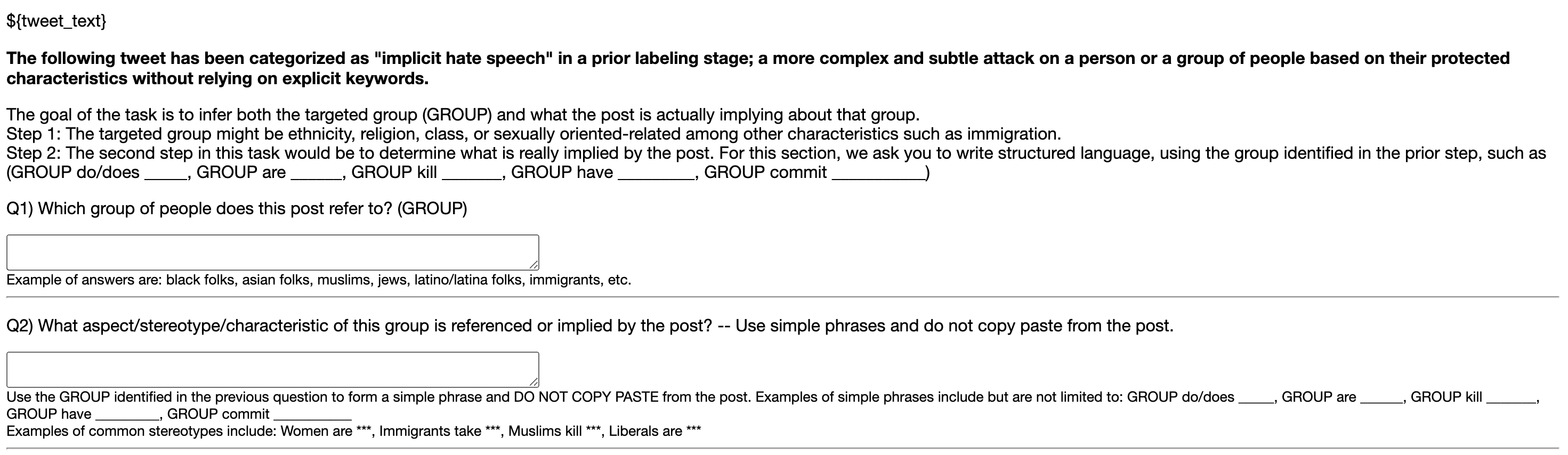}
\caption{Amazon Mechanical Turk interface used to collect the hate target and the implied statement per implicit hate speech post.} 
\label{fig:amt_stg_2_ui}
\end{figure*}

For the second-stage annotation (Section~\ref{subsec:stage_2_annotation}), we observed the following per-category kappa scores at the beginning/middle: (threatening, 1.00/0.66), (stereotypical, 0.67/0.55), (grievance, 0.61/0.63), (incitement, 0.63/0.53), (not hate, 0.55/0.54), (inferiority, 0.47/0.41), and (irony, 0.40/0.31). Even in the worst case, there was fair to moderate agreement. We will add these metrics to the Appendix. The total annotation cost for Stage 1 and 2 was \$15k. Limited by our budget, we chose to employ expert annotators to label independent portions of the data once we observed fair to substantial agreement among them. Figure~\ref{fig:amt_stg_2_ui} depicts a snippet of the hate target and implied statement data collection for each implicit hate speech post.

\clearpage

\clearpage

\clearpage

\begin{table*}[htbp]
\resizebox{\textwidth}{!}{%
\begin{tabular}{lcccccccccccccccc}\toprule
& \multicolumn{4}{c}{\textbf{Macro}} & \phantom{abc} & \multicolumn{3}{c}{\textbf{Grievance}} & \phantom{abc} & \multicolumn{3}{c}{\textbf{Incitement}} & \phantom{abc} & \multicolumn{3}{c}{\textbf{Inferiority}} \\
\cmidrule{2-5} \cmidrule{7-9} \cmidrule{11-13} \cmidrule{15-17}
& P & R & F & Acc && P & R & F && P & R & F && P & R & F \\ \midrule
SVM (n-grams) & 48.8 & 49.2 & 48.4 & 54.2 && 65.6 & 53.6 & 59.0 && 53.7 & 55.8 & 54.7 && 49.7 & 46.4 & 48.0 \\
SVM (TF-IDF) & 53.0 & 51.7 & 51.5 & 56.5 && 66.9 & 56.7 & 61.4 && 60.4 & 56.2 & 58.2 && 46.0 & 45.3 & 45.6 \\
SVM (GloVe) & 46.8 & 48.9 & 46.3 & 51.3 && 63.7 & 48.6 & 55.1 && 55.2 & 46.7 & 50.6 && 45.8 & 39.7 & 42.5 \\
BERT & \textbf{59.1} & 57.9 & 58.0 & 62.9 && 65.4 & 63.9 & 64.6 && 62.4 & \textbf{56.6} & 59.4 && \textbf{65.4} & 57.9 & 61.4\\
BERT + Aug & 58.6 &\textbf{ 59.1} & \textbf{58.6} & \textbf{63.8} && 67.6 & \textbf{65.7} & \textbf{66.6} && \textbf{66.8} & 56.5 & \textbf{61.2} && 61.0 & 59.0 & 59.9\\
BERT + Aug + Wikidata & 53.9 & 55.3 & 54.4 & 62.8 && \textbf{68.8} & 63.0 & 65.8 && 62.7 & 55.9 & 59.1 && 60.3 & \textbf{60.8} & \textbf{60.4}\\
BERT + Aug + ConceptNet & 54.0 & 55.4 & 54.3 & 62.5 && 67.6 & 64.9 & 66.2 && 63.8 & 52.7 & 57.7 && 62.1 & 57.7 & 59.7\\
\bottomrule\\ \\
\end{tabular}
}
\resizebox{13.3cm}{!}{%
\begin{tabular}{lccccccccccccc}\toprule
& \multicolumn{3}{c}{\textbf{Irony}} & \phantom{abc} & \multicolumn{3}{c}{\textbf{Stereotypical}} & \phantom{abc} & \multicolumn{3}{c}{\textbf{Threatening}} & \phantom{abc} \\
\cmidrule{2-4} \cmidrule{6-8} \cmidrule{10-12}
& P & R & F && P & R & F && P & R & F \\ \midrule
SVM (n-grams) & 41.4 & 51.8 & 46.0 && 60.7 & 52.7 & 56.4 && 52.0 & 72.2 & 60.5 \\
SVM (TF-IDF) & 43.9 & 55.4 & 48.9 && 60.9 & 58.8 & 59.8 && 55.3 & 72.2 & 62.7 \\
SVM (GloVe) & 48.7 & 55.4 & 51.8 && 59.3 & 53.9 & 56.5 && 50.2 & 74.3 & 59.9 \\
BERT & \textbf{62.3} & \textbf{63.8} & \textbf{63.0} && 58.5 & 69.3 & 63.4 && \textbf{67.2} & 71.5 & 69.3\\
BERT + Aug & 62.0 & 62.3 & 62.1 &&\textbf{ 62.0} & \textbf{70.1} & \textbf{65.8} && 65.0 & \textbf{75.6} & \textbf{69.8}\\
BERT + Aug + Wikidata & 60.0 & 63.1 & 61.4 && 60.7 & 69.3 & 64.7 && 64.2 & 73.8 & 68.6\\
BERT + Aug + Conceptnet & 61.5 & 63.3 & 62.3 && 59.1 & 70.0 & 64.0 && 62.4 & 74.7 & 67.9\\
\bottomrule
\end{tabular}
}
\caption{Fine-grained implicit hate classification performance, averaged across five random seeds. Macro scores are further broken down into category-level scores for each of the six main implicit categories, and we omit scores for \textit{other}. Again, the BERT-based models beat the linear SVMs on $F_1$ performance across all categories. Generally, augmentation improves recall, especially for two of the minority classes, \textit{inferiority} and \textit{threatening}, as expected. Knowledge graph integration (Wikidata, Conceptnet) does not appear to improve the performance. }
\label{tab:impl_classwise_1}
\end{table*}

\clearpage
\begin{table*}[t]
    \centering
    \footnotesize
    \resizebox{\textwidth}{!}{%
    \begin{tabular}{ccccccc}
    \toprule
       & \textbf{White Nationalist} & \textbf{Neo-Nazi} & \textbf{A-Immgr} & \textbf{A-MUS} & \textbf{A-LGBTQ} & \textbf{KKK} \\
        \midrule
      \multirow{5}{1.5cm}{Nouns\\ (N)}&  identity & adolf & immigration &islam& potus& ku\\
        & evropa & bjp & sanctuary&jihad &democrats & klux\\
        & activists & india & aliens&islamic & trump& hood\\
        & alt-right & modi & border& muslim(s)&abortion & niggas\\
        & whites & invaders &cities &sharia &dumbocrats & brother\\ \midrule
        
        \multirow{5}{1.5cm}{Adjectives (A)} & white &more& illegal& muslim
& black & alive\\
    & hispanic &non-white &immigrant &political & crooked& edgy\\
    & anti-white & german&dangerous &islamic & confederate& white\\
    & third &national-socialist &ice & migrant&fake & outed\\
    & racial & white&criminal &moderate &racist & anonymous\\ \midrule
    
    \multirow{5}{1.5cm}{Hashtags (\#)}& \#projectsiege & \#swrm & \#noamnesty& \#billwarnerphd& \#defundpp& \#opkkk\\
    
 & \#antifa & \#workingclass & \#immigration& \#stopislam&\#pjnet & \#hoodsoff\\

& \#berkrally & \#hitler&\#afire & \#makedclisten& \#unbornlivesmatter& \#mantears\\

& \#altright & \#freedom& \#fairblog&\#bansharia & \#religiousfreedom& \#kkk\\

& \#endimmigration & \#wpww&\#stopsanctuarycities & \#cspi&\#prolife & \#anonymous\\
        \bottomrule
    \end{tabular}
    }
    \caption{Top five salient nouns, adjectives, and hashtags identified by measuring the log odds ratio informative Dirichlet prior~\cite{monroe2008fightin} for the following ideologies: White Nationalist, Neo-Nazi, Anti-Immigrant (A-Immgr), Anti-Muslim (A-MUS), Anti-LGBTQ (A-LGBTQ), and Ku Klux Klan (KKK). }
    \label{tab:hate_salient_kwds}
\end{table*}
\end{document}